\documentclass{article}
\usepackage{amsmath,mlspconf}
\usepackage[utf8]{inputenc} 
\usepackage[T1]{fontenc}    
\usepackage{hyperref}       
\usepackage{url}            
\usepackage{booktabs}       
\usepackage{amsfonts}       
\usepackage{nicefrac}       
\usepackage{microtype}      
\usepackage{lipsum}
\usepackage{graphicx}
\usepackage{xcolor}

\usepackage{flushend}

\usepackage{bbm}

\usepackage[numbers]{natbib}

\usepackage{algorithm}
\usepackage{algorithmic}

\usepackage{color}

\copyrightnotice{978-1-7281-6338-3/21/\$31.00 {\copyright}2021 IEEE}

\toappear{2021 IEEE International Workshop on Machine Learning for Signal Processing, Oct.\ 25--28, 2021, Gold Coast, Australia}

\title{Multimodal data visualization and denoising with integrated diffusion}
\name{Anonymous\thanks{Anonymous.}}
\address{Anonymous}

\name{%
    Manik Kuchroo$^{1,4 *}$%
    \quad Abhinav Godavarthi$^{2 *}$%
    \quad Alexander Tong$^{3}$
    \quad Guy Wolf$^{5 \dagger}$
    \quad Smita Krishnaswamy $^{4,3 \dagger}$\thanks{$^{*}$ Equal contribution. $^{\dagger}$ Equal senior author contribution. This research was partially funded by IVADO Professor funds, CIFAR AI Chair, and NSERC Discovery grant 03267 [\emph{G.W.}]; CZI grants 182702 \& CZF2019-002440 [\emph{S.K.}]; and NIH grants R01GM135929 \& R01GM130847 [\emph{G.W., S.K.}]. The content provided here is solely the responsibility of the authors and does not necessarily represent the official views of the funding agencies. Correspondence to: \texttt{smita.krishnaswamy@yale.edu}}%
}
\address{%
    $^{1}$ Yale University, Dept. of Neuro.;
    $^{2}$ Dept. of Appl. Math.;
    $^{3}$ Dept. of Comp. Sci.;
    $^{4}$ Dept. of Genetics \\%
    $^{5}$ Universit\'{e} de Montr\'{e}al, Dept. of Math. \& Stat.; Mila -- Quebec AI Institute\\%
}

\begin{document}

\maketitle

\begin{abstract}
We propose a method called integrated diffusion for combining multimodal data, gathered via different sensors on the same system, to create a integrated data diffusion operator. As real world data suffers from both local and global noise, we introduce mechanisms to optimally calculate a diffusion operator that reflects the combined information in data by maintaining low frequency eigenvectors of each modality both globally and locally. We show the utility of this integrated operator in denoising and visualizing multimodal toy data as well as multi-omic data generated from blood cells, measuring both gene expression and chromatin accessibility. Our approach better visualizes the geometry of the integrated data and captures known cross-modality associations. More generally, integrated diffusion is broadly applicable to multimodal datasets generated by noisy sensors collected in a variety of fields.
\end{abstract}

\section{Introduction}
\label{sec:intro}

Technological advances have allowed for multimodal instruments to provide information on the same system in parallel. Now, computational approaches must also incorporate the maximum amount of information from all modalities in order to perform a wide variety of downstream tasks, such as integrated visualization, denoising, and cross-modality correlations between features. In the past, solutions have been based on the assumption that naive concatenations of features obtained from unique measurements, or a subset of selected features, can offer viable solutions \cite{Gravina2017,Lahat2015}. Other neural network based approaches have been proposed as well; for instance, domain transfer autoencoders and cycle GANs \cite{zhu2020unpaired}. However these approaches are sensitive to the scale of and noise present in each feature space. This problem is particularly present in high throughput biomedical data, such a single cell RNA-sequencing and single cell ATAC-sequencing, which have entirely different scales and suffer from differing degrees of noise and sparsity. In order to address these concerns, we turn to the framework of data diffusion that was developed by \cite{Coifman2006}.

According to the data diffusion framework, we can learn the intrinsic space of the data by powering a Markov transition matrix to a power $t$, which implicitly calculates a $t$-step random walk on the data graph. This process accumulates probabilities in paths that traverse through relatively dense regions of the data and diminish in sparse outlier regions, inherently denoising the matrix towards predominant axes of variation represented by the low frequency eigenvectors as shown by \cite{van2018recovering}. In \cite{Coifman2006}, the powered diffusion operator is eigendecomposed to uncover intrinsic data dimensions. Since that seminal work, the Markov matrix, also known as a data diffusion operator, has been shown to be useful in many data processing tasks \cite{moon2018manifold}, including denoising \cite{van2018recovering} and dimensionality reduction \cite{Moon19}.

\begin{figure}[t]
\begin{center}
	\includegraphics[width=\linewidth]{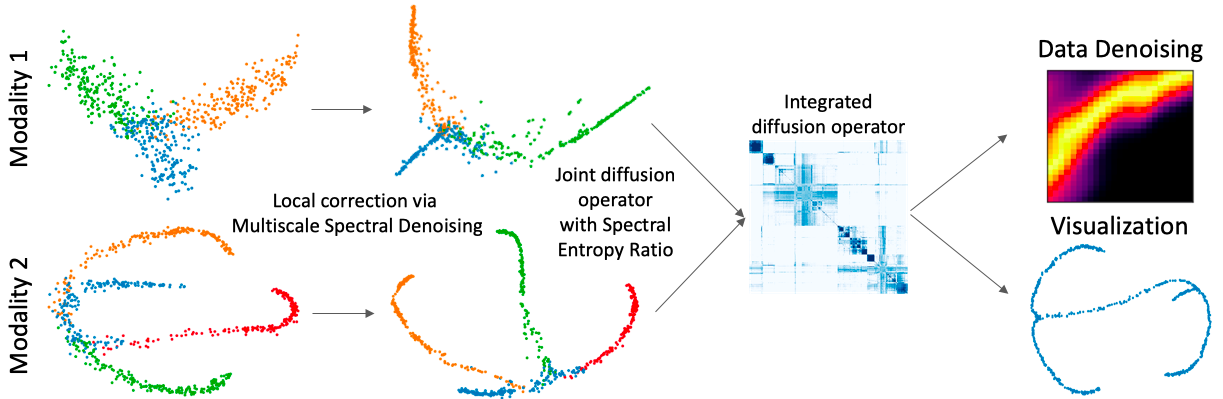}
	\vskip -.1in
	\caption{Workflow of integrated diffusion. We denoise each modality to local low frequency eigenvectors with multiscale spectral denoising. Next, we calculate and compare the intrinsic dimensionality of each dataset via spectral entropy ratio to determine the ideal number of $t$-steps to place in each modality in an alternating random walk. The resulting diffusion operator can help denoise and visualize data.
	\label{fig:overview}}
\vskip -.1in
\end{center}
\end{figure}

Here, we define an \emph{integrated diffusion operator} for multiple data modalities. First, we emphasize dominant directions at a local level in each modality by using a multiscale spectral denoising method to denoise each data modality before modality specific diffuion operator calculation. These diffusion probabilities are then integrated by taking several steps in the data graph from one modality, and several steps on the data graph defined by the other modality. The number of steps is carefully chosen based on \emph{spectral entropy} of each modality. Both of these steps help address modality specific sources of noise both at the local and global levels (Fig. \ref{fig:overview}). Empirically we show that our method yields more accurate visualizations and more faithful denoising on both datasets where ground truth is known and in exploratory biomedical datasets, as compared to a variety of alternative methods for combining multimodal data.

\section{Background}
\label{sec:background}

\subsection{Manifold learning via data diffusion}
Intuitively, while measurement strategies often produce high dimensional observations, their intrinsic dimensionality, or number of degrees of freedom, is relatively low. In ~\cite{Coifman2006}, diffusion maps were proposed as a robust way to capture intrinsic manifold geometry in dataset by eigendecomposing diffusion operators. Using $t$-step random walks that aggregate local affinity, nonlinear relations are revealed in data, which allows their embedding in low dimensions. These local affinities are commonly constructed using a Gaussian kernel
\begin{equation}
\label{equation:GKernel}
\mathbf{K} (x_i, x_j) = \exp\left( {-\frac{\| x_i- x_j\|^2}{\varepsilon}  }\right) \,, \quad i,j=1,...,N
\end{equation}
where $\mathbf{K}$ is an $N \times N$ Gram matrix whose $(i,j)$ entry is denoted by $\mathbf{K}(x_i, x_j)$ to emphasize the dependency on $X$, based on bandwidth parameter $\varepsilon$ that controls local neighborhood sizes. A diffusion operator is defined as the row-stochastic 
\begin{equation}\label{equation:diffusionoperator}
\mathbf{P} = \mathbf{D}^{-1} \mathbf{K} ,
\end{equation}
where $\mathbf{D}$ is diagonal matrix: $\mathbf{D} (x_i, x_i) = \sum_j \mathbf{K} (x_i,x_j)$.

The matrix $\mathbf{P}$, or diffusion operator, defines single-step transition probabilities for a time-homogeneous diffusion process, or a Markovian random walk, over the data. The eigenvectors of $\mathbf{P}$, denoted $\Phi = \phi_0, \phi_1, \ldots, \phi_n$, represent frequency harmonics over the graph based on equivalence to eigenvectors of a normalized graph Laplacian. The eigenvalues $\Lambda = \lambda_1, \lambda_2, \lambda_i \ldots \lambda_n$ directly indicate frequencies, as they are related to the eigenvectors.

The eigenvectors of the  diffusion operator are equivalent to those of the normalized graph Laplacian $L = I-P = D^{-1}L_u=D^{-1}(D-K)$, where $I$ is the identity matrix, $D$ is the degree matrix, $K$ is the kernel affinity, $L_u$ is the unnormalized graph Laplacian. Graph Laplacian eigenvectors have been shown to be equivalent to graph frequency harmonics~\cite{shuman2013emerging}. Thus, signal loadings on to diffusion eigenvectors create a {\em graph Fourier transform} defined as $\Phi^T f$ for a graph signal $f$.

Signals can be filtered using the graph Fourier transform by altering their loading coefficients on to eigenvectors of the graph Laplacian. Thus, a {\em graph filter} can be defined as
\begin{equation}\label{equation:filter}
    h(f) = \Phi h(\Lambda) \Phi^T f
\end{equation}
using a slight abuse of notation with $\Lambda$ being a diagonal matrix of eigenvalues. Here, $h$ rescales the eigenvalues to modulate frequency components of $f$. In~\cite{Coifman2006}, powers $\mathbf{P}^t$ of the diffusion operator, for $t > 0$, not only simulate $t$ step random walks over the data, but can also be seen as soft low-pass graph filters $h(\lambda_i, t) = \lambda_i^t$, which diminish higher frequency noise components more rapidly than lower frequency informative components. In~\cite{van2018recovering}, such filters were used on biological data to denoise single cell RNA sequencing measurements by simply applying the powered diffusion operator (or {\em diffusion filter}) to the data as
\begin{equation} \label{equation:denoising}
\hat{X} = \mathbf{P}_X X ,
\end{equation}
thus avoiding eigendecomposition.

\subsection{Alternating diffusion}
Recently, alternating diffusion has been proposed to combine diffusion operators created from multimodal data \cite{Katz2019}. Intuitively, this generalizes the random walk to ``hop'' between different metric spaces by taking a matrix product of the Markov transition matrices
\begin{equation}
\label{Kernel Product}
    \mathbf{P} (x_i, x_j) = \mathbf{P_{i}} \mathbf{P_{j}} \
\end{equation}
Finally, the resultant alternating diffusion operator $\mathbf{P}$ is powered to stimulate ``hopping'' across modalities. While this approach is able to construct a joint diffusion operator, it is sensitive to local and global noise found in each dataset creating a joint manifold that represents not only modality specific sources of information, but also noise.

\section{Method}

\begin{figure*}[t]
\begin{center}
	\includegraphics[width=.9\linewidth]{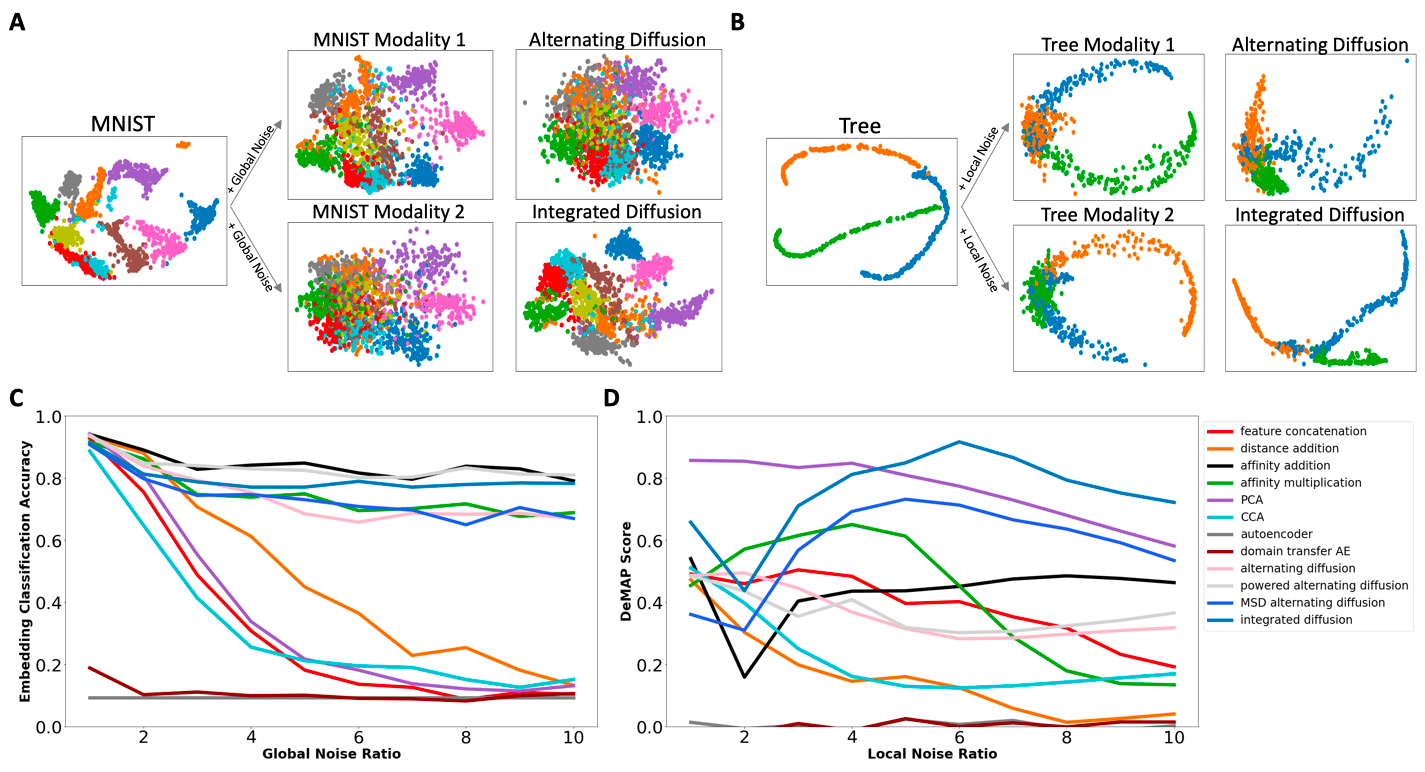}
	\vskip -.2in
	\caption{Generation of multimodal data from ground truth MNIST and artificial tree data as well as visualization comparisons. A) Gaussian noise is added to the entire MNIST datasets to simulate global noise as visualized by PHATE \cite{Moon19}. B) Gaussian noise is added to branches of the artificial tree dataset to simulate local noise as visualized by PHATE. C) Classification accuracy for predicting MNIST digits from integrated embeddings created by a variety of techniques with increasing differences in global noise. D) DeMAP correlations between ground truth tree distances with integrated embedding distances created by a variety of techniques with increasing differences in local noise.
	\label{fig:viz}}
\end{center}
\end{figure*}

\subsection{Problem formulation and approach}

Let $\mathbf{X} \subseteq \mathbbm{R}^{D_X}$ and $\mathbf{Y} \subseteq \mathbbm{R}^{D_Y}$ be two datasets generated by measuring the same system with two sets of sensors in different metric spaces. Each of these datasets contains nonoverlapping features with different scales, and is subject to differing degrees of noise. Here, we propose an approach for generating an integrated diffusion operator that selects information from both modalities at {\em multiple scales}. Our approach is based on the idea that frequency components of the diffusion operator can be low-pass filtered using particular powers of this operators. By using multiple scales, we perform both local and global denoising of the data modalities to a degree where highly relevant information is retained in the joint integrated operator. This integrated diffusion operator can then be used to visualize and denoise both datasets (Fig. \ref{fig:overview}).

\subsection{Neighborhood reconstruction via multiscale spectral denoising}
\label{sec:localdenoise}

Specific areas of each modality's data manifold can contain different amounts of noise that may obscure structure in joint embeddings. Therefore, we translate the global denoising idea from \cite{van2018recovering} to local regions and multiple scales by creating and applying hierarchical sets of diffusion operators as described in Alg.~\ref{alg:mgd}. This recursive approach calculates increasingly local diffusion operators and denoises the original modality-specific data at multiple scales. At each scale, the original input data, represented in by $X$ (see Alg.~\ref{alg:mgd}), is averaged with the denoised data $\hat{X}$. Each scale of the hierarchy contributes half as much correction as the previous scale, with the overall effect summing to one. We apply this modality specific local denoising approach to correct all data modalities before integrating them. It should be noted that the filter we use from Eq.~\ref{equation:denoising} can be replaced with a more general filter, i.e., any filter that takes the form of Eq.~\ref{equation:filter}. 

\begin{algorithm}[tb]\label{alg:mgd}
    \caption{Multiscale Graph Denoising $\texttt{MGD}(X, t, \tau, c)$}
    \label{alg:id}
\begin{algorithmic}
    \STATE {\bfseries Input:} dataset $X$, local denoising timestep $t$, a minimum cluster size $\tau$, and number of clusters $c$.
    \STATE {\bfseries Output:} an approximate and locally dataset $\hat{X}$.
    \IF {$ |X| < \tau$} 
        \STATE Return $X$
    \ENDIF
    \STATE Compute diffusion operator $P$ as described by equations \ref{equation:GKernel} and \ref{equation:diffusionoperator} 
    \STATE $\hat{X} = P^t X$ as described in equation \ref{equation:denoising}
    \STATE $C_1, C_2, \ldots, C_c = \texttt{SpectralCluster}(P)$
    \STATE Return $(X + \bigcup_{i = 1}^c \texttt{MGD}(\hat{X}[C_c], t, \tau, c)) / 2$
\end{algorithmic}
\end{algorithm}

\subsection{Global denoising via spectral entropy}
\label{sec:globalnoise}

In addition to correcting for varying local noise within a single modality, it is crucial to only maintain the most globally important eigenvectors in each data modality. While \cite{van2018recovering} did globally denoise by taking the diffusion operator powers $P^t$, the methodology used there for tuning $t$ essentially required manual trial and error. Such tuning is crucial here as a small $t$ could incorporate significant modality specific noise, while a high $t$ could improperly diminish the effect of informative eigenvectors. Therefore, we propose to select $t$ for each data modality separately by using {\em spectral entropy} to evaluate how much information it encodes in the diffusion operator for each candidate value of $t$. We can then methodically tune $t$ to be a scale where the information loss stabilizes, with the reasoning that signals are harder to remove than the noise. 

Spectral entropy is defined as the Shannon entropy of normalized eigenvalues, i.e., 
\begin{equation} \label{equation:spectral_entropy}
S(P, t) = - \sum_i \psi_i^t \log(\psi_i^t),
\end{equation}
which in this context quantifies the spread of information throughout the eigenspectrum of the diffusion operator. We reason that innumerable noise dimensions will quickly drop off while informative dimensions are harder to remove. Hence, we choose the elbow of this curve to find an inflection point $k$ in the spectral entropy $S(p)$ as can be seen in Fig.~\ref{fig:bio}C.

\subsection{Fusion of operators}
While the spectral entropy heuristic is used to compute $t$ for each modality independently, it cannot directly be used to tune timescales across modalities to combine their diffusion operators together. Since noise may be inherently present in the system being measured, and therefore be present in both datasets, the integrated operator must again be used by taking its powers to a given time scale. Powering directly by $\mathbf{t_{1}}$, $\mathbf{t_{2}}$ and $\mathbf{t_{integrated}}$ would lead to an oversmoothing effect that would eliminate information from the low frequency eigenvectors in the final computed manifold, effectively collapsing independent informative data points together. To alleviate this concern, we raise each modalities diffusion operator to the lowest possible multiple of the ideal view-specific $t$. This means we can write our integrated diffusion operator, $\mathbf{J}$, to reflect the differing levels of global information between views as
\begin{equation}
    \mathbf{J} = \mathbf{P_{1}^{t_{1}} \mathbf{P_{2}^{t_{2}}}}
\end{equation}
where $\mathbf{t_{1}}$ and $\mathbf{t_{2}}$ are integer values obtained from the reduced ratio as described above, and $\mathbf{P_{1}}$ and $\mathbf{P_{2}}$ are modality specific diffusion operators. This integrated diffusion operator can be applied directly to one of the data modalities as a low pass denoising diffusion filter as done in Eq.~\ref{equation:denoising} or can be powered and embedded using the methods of~\cite{Coifman2006} or \cite{Moon19}.

\begin{figure}
	\includegraphics[width=0.95\linewidth]{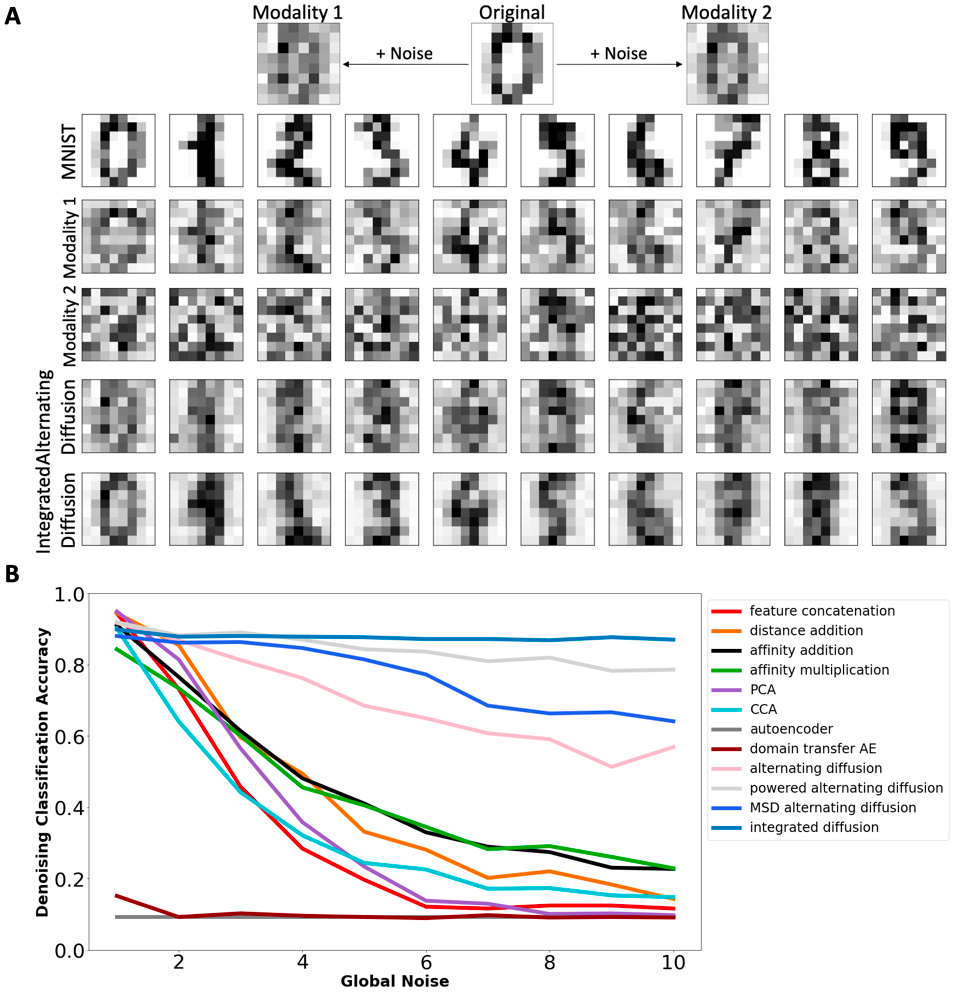}
	\vspace{-0.1in}
	\caption{A) Denoising of MNIST data with alternating (middle row) and integrated (bottom row) diffusion operators. B) MNIST denoised pixel classification accuracy with differing diffusion-based multimodal denoising strategies. As above, data was generated by adding differing amounts of random gaussian noise to MNIST pixels.}
\label{fig:denoise}
\end{figure}

\section{Experimental Results}
In the following experiments we evaluate the ability of integrated diffusion to visualize and denoise high-dimensional multimodal data. We first simulate global noise using the MNIST handwritten digits dataset.  We generate multiple modalities of MNIST handwritten digits by adding Gaussian noise to the images, where each pixel value $p_i' = p_i +N(\nu)$, where $\nu$ changes based on the level of noise. To showcase the ability of our method to handle modalities with significant differences in global noise, we add a fixed amount of Gaussian noise to simulate one data modality and increasing amount of Gaussian noise to simulate the second data modality (Fig. \ref{fig:viz}A). Next, we generate multiple modalities of high dimensional artificial trees with varying amounts of local noise to specific branches. Similar to our MNIST multimodal datasets, we add a small amount of fixed noise to each tree before adding increasing amounts of noise to differing branches (Fig. \ref{fig:viz}B). Finally, we apply our integrated diffusion approach to real world single cell biological data from {RNA}-sequencing (gene expression) and ATAC-sequencing (chromatin accessibility). With these datasets, we compare integrated diffusion to other multimodal learning approaches on visualization and denoising tasks.

\subsection{Visualization}
To quantify the differences in visualizations produced from differing multimodal integration strategies, we compared the first 20 diffusion map components computed from diffusion operator constructions based off of multimodal feature concatenation, distance addition, affinity addition, affinity multiplication and alternating diffusion, to integrated diffusion. We also performed ablation studies, comparing these techniques to various diffusion operators: alternating diffusion with local correction and alternating diffusion with modality specific powering of diffusion operators via spectral entropy ratio. We also compared to non-diffusion based embeddings produced by cycle GANs, autoencoders and domain transfer autoencoders. For our MNIST comparisons, we train a kNN-classifier to predict MNIST digit of origin from the embedding trained from each technique. For our tree comparisons, with branches with differing amounts of local noise, we try to determine how successful our embeddings are using DeMAP (Denoised Manifold-Affinity Preservation) proposed in \cite{Moon19}. DeMAP computes geodesic distance between all data points in a noiseless dataset and correlates it with distance between these data points in an embedding. This method tries to determine if the embedding accurately maintains ground truth point to point distances in compressed space.

All strategies performed comparably when both modalities had a similar degree of local and global noise. As the difference in global noise increased in our MNIST embedding classification task, strategies that powered the diffusion operators to account for global noise outperformed strategies that did not (Fig. \ref{fig:viz}C). When embedding trees with varying degrees of local branch specific noise, methods that perform local correction with multiscale spectral denoising significantly outperformed methods that did not (Fig. \ref{fig:viz}D).

\subsection{Denoising}
Previous work in diffusion filters has shown that low pass filtering can correct many types of noise present in real world datasets, allowing for downstream analysis \cite{van2018recovering}. Here, we compare several methods for data denoising with our integrated diffusion approach. As done previously, we created multimodal MNIST data by adding differing amounts of global noise. After computing the integrated diffusion operator with each of these comparison methods, we filter the noisier MNIST modality as done previously \cite{van2018recovering} and as can be seen in Fig. \ref{fig:denoise}A. To get quantitative results, we train a kNN-classifier on the denoised pixel values to determine how well each operator is able to predict the digit (Fig.~\ref{fig:denoise}). As shown in Fig.~\ref{fig:denoise}B, across all denoising comparisons, classification accuracy on increasingly noisy MNIST digits were best recovered by integrated diffusion followed by alternating diffusion with modality specific view powering, both methods account for global information within each noisy modality.

\begin{figure*}[!t]
\begin{center}
	\includegraphics[width=0.92\linewidth]{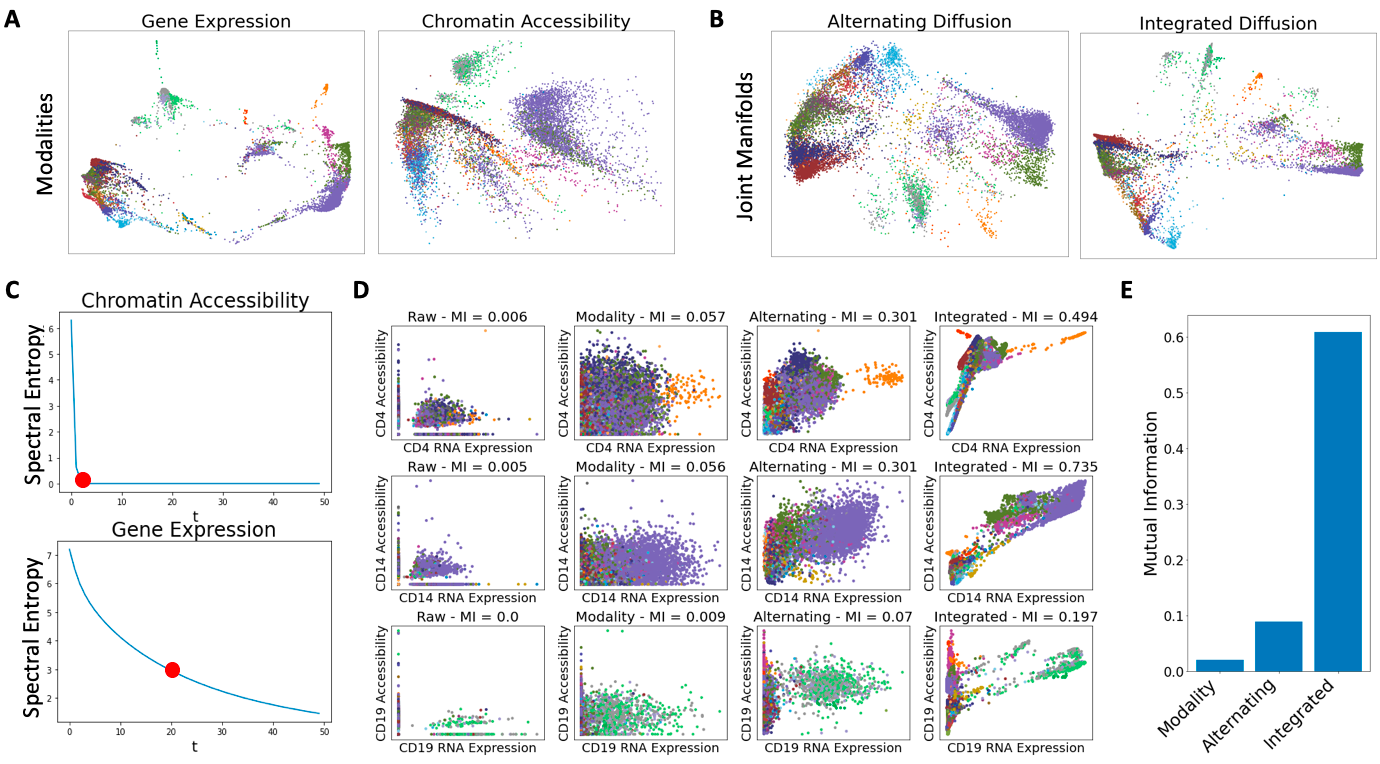}
	\caption{Application of integrated diffusion to multimodal single cell data. A) Visualization of gene expression and chromatin accessibility manifolds as well as B) alternating and integrated manifolds via PHATE. Points colored by annotated cell type. C) Visualization of spectral entropy of each modality. D) Mutual information between the expression of a gene and it's accessibility across differing denoising strategies. E) Average mutual information for differing denoising strategies across all gene expression-gene accessibility pairs.
	\label{fig:bio}}
\end{center}
\vskip -0.2in
\end{figure*}

\section{Biological Applications}

New methods allow for the measurement of tens to hundreds of thousands of features in single cells, allowing for unprecedented insight into biological and cell type specific processes. Until recently, only a single modality could be measured in each cell, be it expression of genes through RNA sequencing or the accessibility of chromatin regions through ATAC sequencing. Now novel techniques allow for the measurement of different modalities at single-cell resolution. Increasingly commonly, individual cells are measured with a combination of chromatin accessibility and RNA expression \cite{Ma2020, Cao2018}. This new type of data is powerful, as it not only allows for the study of each modality independently, but also allows for the discovery of regulatory mechanisms between modalities. Currently, no computational techniques are capable of modelling and predicting these dynamics as there are no strategies that integrate different modalities of data to jointly visualize and denoise multimodal single-cell data.

We apply integrated diffusion to multimodal single cell data of 11,909 blood cells, visualizing the integrated manifold and uncovering key cross modalities interactions. Visualizing each modality, gene expression and chromatin accessibility, independently reveals similar overall structure, but different resolutions. Chromatin accessibility data, when compared to gene expression data, is incredibly sparse and generally considered to be far more noisy. When computing the spectral entropy of each modality, we can clearly see that the chromatin accessibility diffusion operator has a far fewer informative dimensions than the gene expression operator. The alternating diffusion approach, which does not account for modality specific sources of noise creates an embedding that blends the distinct structure of gene expression data with the less informative structure of chromatin accessibility data. Integrated diffusion, however, appears to better resolve differences in information across datasets, producing a visualization that contains sharper borders between populations and displays clear structure when visualized with PHATE (Fig.~\ref{fig:bio}A-B).

A major issue in single cell data is sparsity, which makes it very difficult to measure and model cross modality interactions. Theoretically, if a gene is expressed, then the chromatin encoding that gene must be accessible. With this understanding of the data, we try to recover these known associations between gene expression and chromatin accessibility (Fig.~\ref{fig:bio}D). Due to sparsity, there is no association as computed by mutual information between these variables without denoising with Eq.~\ref{equation:denoising}. There are several strategies to recover these cross modality interactions: denoising each modality with modality-specific diffusion operators, denoising with a single alternating diffusion operator or denoising with a single integrated diffusion operator. Using the integrated diffusion operator appears to best recover known gene expression and chromatin accessibility associations as shown in genes CD19, CD14 and CD4 (Fig.~\ref{fig:bio}D).  We then computed these associations across all genes with each of our denoising strategies. Across 18,659 genes, integrated diffusion recovered significantly more information between a gene's accessibility and its expression than alternating diffusion and modality-specific diffusion (Fig. \ref{fig:bio}E).

\section{Conclusion}

We introduce the integrated diffusion operator for learning an integrated data geometry as described by multiple data measurement modalities applied to a single system. We show its improvement over more naive methods on synthetic and biological datasets. We apply our method in biomedical setting to a multiomic dataset, where we generated rich integrated manifolds and recover cross modality gene-chromatin associations. Our flexible framework is extendable to multiple modalities and we expect it will allow for successful integration and analysis of massive datasets in a wide variety of fields.

\bibliography{refs,strings}
\bibliographystyle{IEEEbib}

\end{document}